\documentclass{article}
\usepackage{ijcai25}

\usepackage{times}
\usepackage{soul}
\usepackage{url}
\usepackage[hidelinks,colorlinks=true, linkcolor=blue, citecolor=black, urlcolor=blue]{hyperref}
\usepackage[utf8]{inputenc}
\usepackage[small]{caption}
\usepackage{graphicx}
\usepackage{amsmath}
\usepackage{amsthm}
\usepackage{booktabs}
\usepackage{algorithm}
\usepackage{algorithmic}
\usepackage[switch]{lineno}

\title{DejAIvu: Identifying and Explaining AI Art on the Web in Real-Time with Saliency Maps}

\author{
    Jocelyn Dzuong
    \affiliations
    Florida International University
    \emails
    jdzuo001@fiu.edu
}

\begin{document}

\maketitle

\begin{abstract}
The recent surge in advanced generative models, such as diffusion models and generative adversarial networks (GANs), has led to an alarming rise in AI-generated images across various domains on the web. While such technologies offer benefits such as democratizing artistic creation, they also pose challenges in misinformation, digital forgery, and authenticity verification. Additionally, the uncredited use of AI-generated images in media and marketing has sparked significant backlash from online communities. In response to this, we introduce DejAIvu, a Chrome Web extension that combines real-time AI-generated image detection with saliency-based explainability while users browse the web. Using an ONNX-optimized deep learning model, DejAIvu automatically analyzes images on websites such as Google Images, identifies AI-generated content using model inference, and overlays a saliency heatmap to highlight AI-related artifacts. Our approach integrates efficient in-browser inference, gradient-based saliency analysis, and a seamless user experience, ensuring that AI detection is both transparent and interpretable. We also evaluate DejAIvu across multiple pretrained architectures and benchmark datasets, demonstrating high accuracy and low latency, making it a practical and deployable tool for enhancing AI image accountability. The code for this system can be found at \href{https://github.com/Noodulz/dejAIvu}{https://github.com/Noodulz/dejAIvu}.
\end{abstract}

\section{Introduction}
Generative AI models such as diffusion models \cite{sohlDeep} and GANs \cite{goodfellowGANs} have reshaped the digital landscape by enabling seamless, high-quality image generation. However, their widespread accessibility raises critical concerns about attribution, authenticity, and ethical use. \cite{hagendorffEthics,zhangDeepfakeAttribution}. Such unauthorized or uncredited use of AI-generated images in media, marketing, and online platforms has led to widespread misinformation, digital forgery, and artist displacement. Moreover, there is growing public demand for increased accountability and transparency regarding the use of AI-generated images \cite{solaimanSocialImpact,shaikhRiseCreative}.

Although there are many AI image detection methods \cite{guoImageDetection,liAdversarial,zhongTexturePatch}, they are often hampered by real-world applicability or lack thereof. Additionally, most detection models operate as black-box classifiers, providing no explainability on why an image is considered AI-generated \cite{yanSanityCheck,gaintsevaImproving,aiornot}. This lack of interpretability reduces trust in detection methods, further exacerbating concerns over misinformation and algorithmic opacity.

We introduce DejAIvu, a browser-based extension enabling real-time detection of AI-generated images while providing saliency-based explainability to enhance transparency and interpretability \cite{simonyanSaliency}. The extension allows for automatic analysis of images on popular image sharing sites without requiring manual uploads or additional effort. DejAIvu leverages an ONNX-optimized deep learning model, which is converted from a TensorFlow-trained convolutional neural network (CNN) to enable efficient in-browser inference \cite{jinONNX}. By running locally within the Chrome extension, it eliminates the need for server-side computation, ensuring low-latency performance and preserving user privacy \cite{wangInferenceBrowsers,hidaka2024distml}. The model has been trained on a curated dataset of more than 270,000 human and AI-generated artworks, incorporating various styles and sources to enhance detection accuracy \cite{elkholy_ai_artwork_attention,elkholy_human_ai_artwork}. Beyond classification, DejAIvu introduces explainability by generating gradient-based saliency maps \cite{smilkovSmoothgrad,selvarajuGradcam,bachLRP,sundarajanAxiomatic}, which highlight regions of an image that contribute most to the model’s classification decision. By visualizing these areas, users can better understand which image features influence AI-generated detection.

\begin{figure*}[t]
    \centering
    \includegraphics[width=0.9\textwidth]{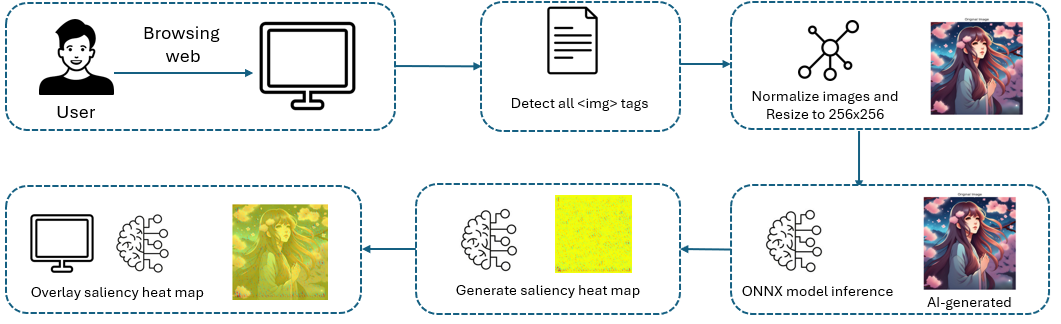}
    \caption{Overview of DejAIvu Pipeline.}
    \label{fig:pipeline}
\end{figure*}

\section{DejAIvu Overview}
\subsection{Architecture and Workflow}

DejAIvu operates by dynamically intercepting images on webpages, passing them through an ONNX-optimized AI model for classification, and generating saliency maps to highlight AI-specific artifacts \cite{jarrett2025gradient,simonyanSaliency}. As demonstrated in Figure~\ref{fig:pipeline}, the system follows the following pipeline:

\begin{enumerate}
    \item \textbf{Image Interception}: The content script identifies images (\texttt{<img>} elements) dynamically as a user browses the web.
    \item \textbf{Preprocessing \& Model Inference}: The image is resized to $256\times256$ pixels, normalized, and fed into the ONNX model for classification.
    \item \textbf{Saliency Heatmap Generation}: If the model detects an image as AI-generated, a gradient-based saliency map is computed.
    \item \textbf{Overlay Visualization}: The saliency map is blended with the original image in real time, allowing users to visually interpret AI artifacts.
\end{enumerate}

\subsection{Model Training \& Optimization}
\begin{table}[h]
    \centering
    \renewcommand{\arraystretch}{1.0} 
    \setlength{\tabcolsep}{3pt}  
    \small  
    \begin{tabular}{|p{2.8cm}|c|c|c|c|}
        \hline
        \textbf{Original Dataset} & \textbf{\# Images} & \textbf{Class} & \textbf{Dim.} & \textbf{Format} \\
        \hline
        DiffusionDB Large  & 39,322  & AI  & 512×512  & .png  \\
        DiffusionDB Small  & 26,212  & AI  & 224×224  & .jpeg  \\
        AI-ArtBench LD     & 62,092  & AI  & 256×256  & .jpeg  \\
        AI-ArtBench SD     & 62,923  & AI  & 768×768  & .jpeg  \\
        WikiArt            & 81,444  & Human  & Variable  & .jpeg  \\
        \hline
    \end{tabular}
    \caption*{Table 1: Datasets used for AI-generated and human-created art classification from \cite{elkholy_ai_artwork_attention}}
    \label{tab:dataset}
\end{table}

\textbf{Computational Resources} The DejAIvu tool was trained using a high-performance computing setup consisting of two NVIDIA RTX 6000 Ada GPUs (each with 48GB VRAM) and 1TB of system RAM. The training pipeline leveraged CUDA 12.2 and was optimized with mixed-precision training to maximize memory efficiency. Training was conducted on high-resolution images (256×256 pixels), with gradient accumulation and automatic mixed precision (AMP) enabled to accelerate convergence \cite{micikevicius2018mixedprecisiontraining,jia2018highlyscalabledeeplearning}.

\textbf{Datasets} As seen in Table~\ref{tab:dataset}, the underlying AI model for DejAIvu is trained on a curated dataset of 271,993 AI-generated and human-created artworks sourced from \cite{elkholy_ai_artwork_attention} via combined datasets from Kaggle and other open source datasets across the web \cite{wangDiffusionDB,liaoArtBench,wikiart}. This dataset is highly imbalanced, with a majority of samples being AI-generated images. To address this imbalance, we apply log-based bias initialization to prevent the model from disproportionately favoring the dominant class \cite{karpathy2019recipe,haixiang2017learning}. Specifically, we initialize the final classification layer using a log-ratio bias adjustment, computed as:

\begin{equation}
b = \log\left(\frac{\text{AI-Generated Samples}}{\text{Human-Made Samples}}\right) = \log\left(\frac{190,549}{81,457}\right)
\end{equation}

This bias is then incorporated into the model, setting its initial predictions to align with the dataset's actual distribution.

\textbf{Dataset Splits} To ensure robust model generalization, the dataset is split into training, validation, and test sets following a stratified partitioning strategy. Specifically, the training set comprises 60\% of the available dataset and is augmented with transformations such as horizontal flipping, slight rotations, and contrast adjustments to mitigate overfitting. The remaining 40\% of the dataset is allocated to validation and test sets. To further ensure a reliable evaluation protocol, the validation and test sets are equally split from this subset, with each receiving 20\% of the original dataset. This results in a final 60\%-20\%-20\% training-validation-test distribution.

\textbf{Preprocessing Pipeline}
The preprocessing pipeline ensures that input images are properly prepared before being fed into the model. First, all images are resized to a fixed dimension of 256 × 256 pixels to maintain consistency across inputs. To ensure numerical stability during inference, pixel values are normalized by scaling them to the range [0,1]. Additionally, data augmentation techniques are applied to improve the model’s generalization and robustness \cite{XuSurvey}. These augmentations include horizontal flipping, slight rotations, and contrast adjustments, which help prevent overfitting and enhance the model’s ability to handle variations in input data \cite{shorten2019survey}. Finally, the trained model is converted to the ONNX format, allowing efficient execution directly in the browser using ONNX.js, which optimizes performance and compatibility with web-based applications.

\section{Experimental Results \& Performance}

To assess the performance of DejAIvu's AI detection model, we compare it against various state-of-the-art architectures, evaluating accuracy, precision, recall, loss, and model size. With this we aim to optimize detection accuracy while ensuring real-time efficiency in the browser.

\paragraph{Model Performance Comparison}  
Table~\ref{tab:model_comparison} summarizes key metrics for multiple architectures. ResNet50 \cite{he2015deep} achieves the highest accuracy of 97.1\% with a reasonable file size of 90.6MB, an improvement from the model of \cite{elkholy_ai_artwork_attention} which attained an accuracy of 95\% with 163.8MB. Utilizing a vision transformer \cite{dosovitskiy2020image} also demonstrates competitive performance (96.0\%) at an exceptionally small size of 1.7MB in contrast to ResNet50 and \cite{elkholy_ai_artwork_attention}'s models. While lighter models like MobileNetV2 (9.2MB) and DenseNet121 (28.3MB) \cite{sandler2018mobilenetv2,huang2017densely} offer reasonable performance, they fall short in precision and recall compared to larger models.

\begin{table}[h]
    \centering
    \small
    \setlength{\tabcolsep}{2pt}  
    \renewcommand{\arraystretch}{1.1}  
    \begin{tabular}{|l|c|c|c|c|c|}
        \hline
        \textbf{Model} & \textbf{Acc.} & \textbf{Loss} & \textbf{Prec.} & \textbf{Rec.} & \textbf{Size (MB)} \\
        \hline
        \textbf{ResNet50} & \textbf{0.971} & \textbf{0.081} & \textbf{0.981} & \textbf{0.978} & \textbf{90.6} \\
        VisionTransformer & 0.960 & 0.113 & 0.968 & 0.974 & 1.7 \\
        ElKholy\_model & 0.950 & 0.145 & 0.970 & 0.958 & 163.8 \\
        VGG16 & 0.929 & 0.190 & 0.941 & 0.958 & 56.2 \\
        DenseNet121 & 0.919 & 0.214 & 0.928 & 0.959 & 28.3 \\
        MobileNetV2 & 0.882 & 0.284 & 0.897 & 0.939 & 9.2 \\
        Xception & 0.782 & 0.586 & 0.921 & 0.754 & 80.1 \\
        NASNetMobile & 0.767 & 0.496 & 0.826 & 0.846 & 18.9 \\
        \hline
    \end{tabular}
    \caption{Performance comparison after training with standard and custom architectures.}
    \label{tab:model_comparison}
\end{table}

\textbf{Latency Comparisons} ResNet-50 was selected as the backbone model for DejAIvu due to its strong balance between accuracy and computational efficiency. To ensure real-time performance, conversion to ONNX.js was chosen over TensorFlow.js (TF.js) for web-based inference. Table~\ref{tab:resnet50_latency} shows that by leveraging ONNX.js instead of TensorFlow.js, DejAIvu achieves an approximate 35ms reduction in latency per image on average, significantly improving real-time inference speed while also preserving accuracy \cite{wangInferenceBrowsers}.

\begin{table}[h]
    \centering
    \begin{tabular}{l c}
        \hline
        \textbf{Model} & \textbf{Avg. Latency} \\
        \hline
        ResNet-50 (ONNX.js) & ~95ms \\
        ResNet-50 (TF.js) & ~130ms \\
        \hline
    \end{tabular}
    \caption{Performance Benchmarks for ResNet-50}
    \label{tab:resnet50_latency}
\end{table}

\section{Demonstration}

To illustrate the real-world applicability of DejAIvu, we demonstrate its functionality across Google Images, highlighting its ability to seamlessly activate with search results. The extension runs in the background, intercepting image requests, detecting AI-generated content, and overlaying explainability heatmaps in real time.

\begin{figure}[h]
    \centering
    \includegraphics[width=\columnwidth]{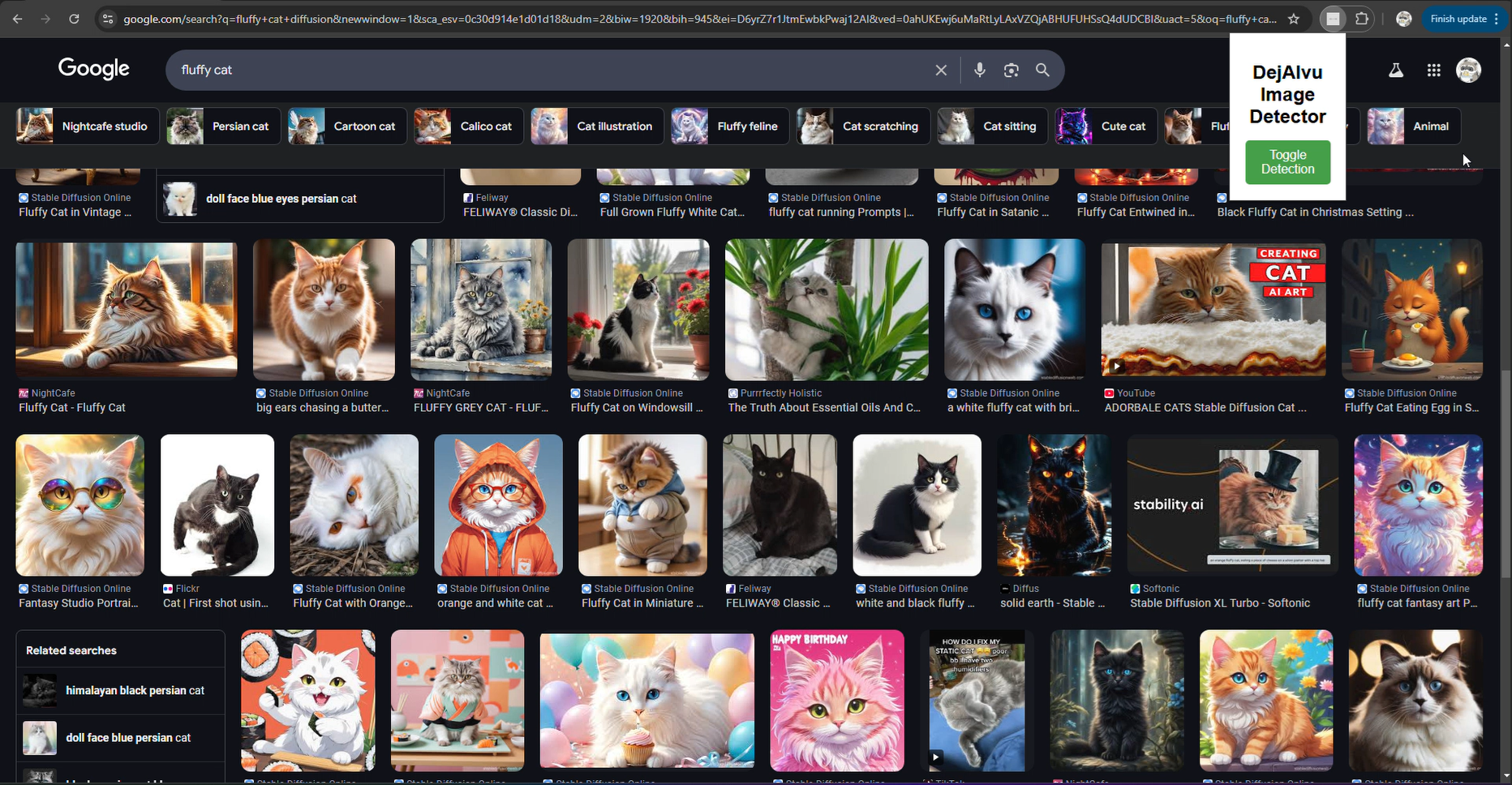}
    \caption{Google Images results of "fluffy cat" before toggling detection}
    \label{fig:google_demo}
\end{figure}

\begin{figure}[h]
    \centering
    \includegraphics[width=\columnwidth]{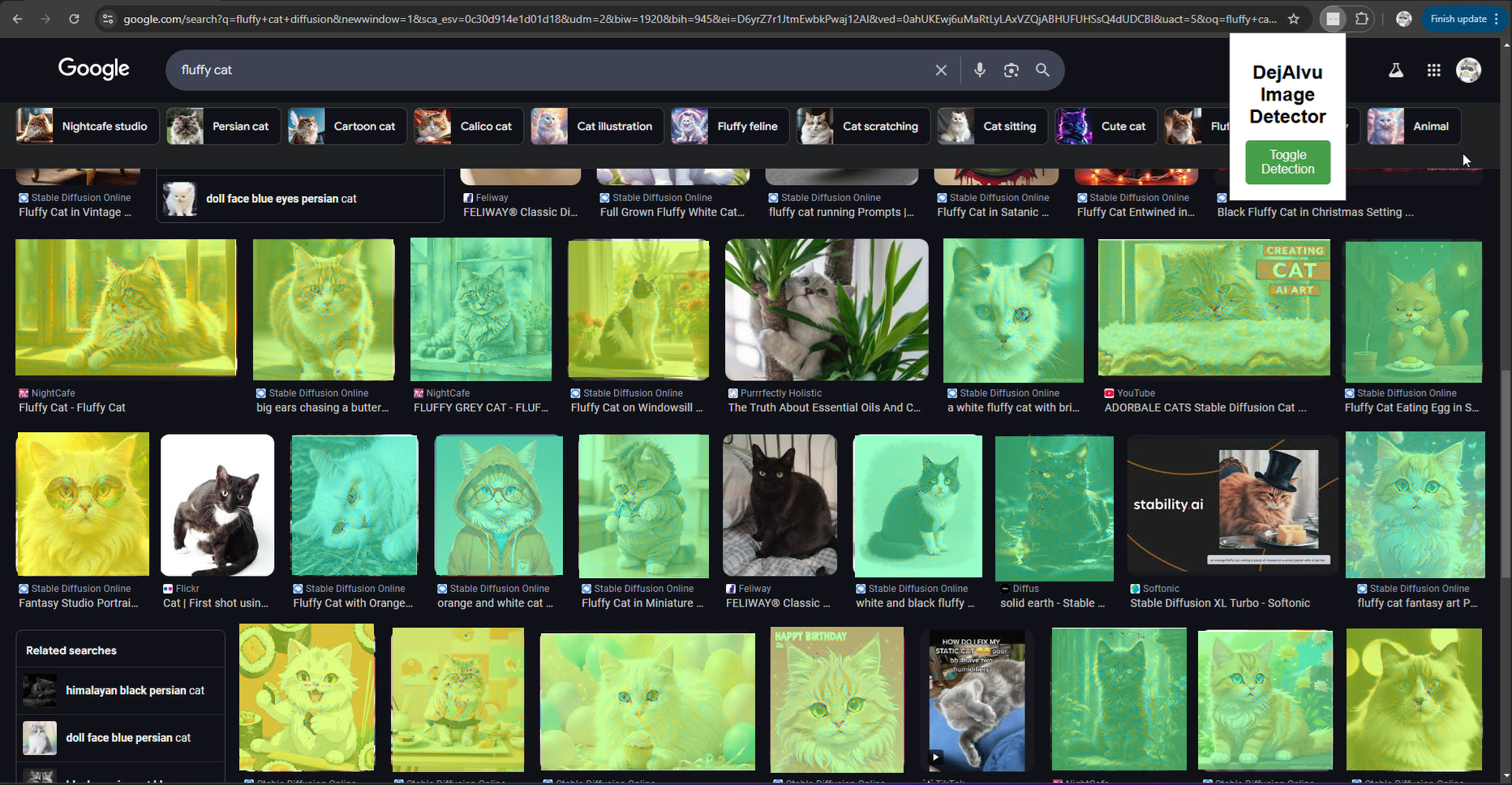}
    \caption{Google Images results of "fluffy cat" after toggling detection}
    \label{fig:pinterest_demo}
\end{figure}

\subsection{User Interaction Flow}
The user experience with DejAIvu is designed to be intuitive and requires minimal setup. Upon installation, the extension becomes active in the background. When the user navigates to an image-heavy website, DejAIvu automatically detects all displayed images, analyzes them in real time, and overlays a Vanilla Gradient heat map on each image inferred as AI generated. Figure~\ref{fig:google_demo} and Figure ~\ref{fig:pinterest_demo} demonstrate the usage and overlay of Vanilla Gradients upon toggling the extension on a Google Images result when the user looks up cats.

\section{Conclusion and Future Work}

DejAIvu introduces a novel approach to AI-generated image detection by providing a fully browser-integrated solution with real-time inference and saliency-based explainability. Unlike existing detection methods that rely on manual uploads or cloud-based inference, DejAIvu operates locally using ONNX.js, ensuring low-latency performance and enhanced user privacy. Through its integration with online platforms such as Google Images, it enables users to transparently identify AI-generated content without disrupting their browsing experience. Our results show high classification accuracy while maintaining computational efficiency. Future work will focus on optimizing inference speed, quantifying latency across image sets, conducting large-scale user testing, and expanding dataset coverage to improve generalization across diverse art styles. Additional features such as adjustable detection thresholds, heat map customization, and per-site detection preferences are also planned to enhance user experience and flexibility.

\bibliographystyle{named}
\bibliography{ijcai25}

\end{document}